\documentclass[letterpaper, 10 pt, conference]{ieeeconf}
\IEEEoverridecommandlockouts
\usepackage{graphics}
\usepackage{algorithm}
\usepackage{algpseudocode}

\usepackage{graphicx}
\usepackage{amsmath}
\usepackage{amssymb}
\usepackage{soul}
\usepackage{subcaption}
\usepackage{wrapfig}
\usepackage{float}
\usepackage{multirow}
\usepackage{textcomp}
\usepackage{lipsum}
\usepackage[english]{babel}
\usepackage[autostyle]{csquotes}
\usepackage[colorinlistoftodos]{todonotes}

\title{\LARGE \bf
Surface Disinfection using Ultraviolet Light\\ with a Mobile Manipulation Robot
}

\author{Alan G. Sanchez$^{1}$ and William D. Smart$^{1}$%

\thanks{This work was partially funded by the National Institutes of Health, under award R01-EB024330.}
\thanks{$^{1}$Alan G. Sanchez and William D. Smart are with the Collaborative Robotics and Intelligent Systems (CoRIS) Institute,
    Oregon State University, Corvallis OR.
    {\tt\small sanchala@oregonstate.edu}, {\tt\small smartw@oregonstate.edu}}%
}

\begin{document}
\maketitle
\thispagestyle{empty}
\pagestyle{empty}

\begin{abstract}
    Robots are being increasingly used in the fight against highly-infectious diseases such as Ebola, MERS, and SARS-COV-2.  Many of the robots that are being used employ ultraviolet lights mounted on a mobile base to inactivate the pathogens.  However, these lights are often mounted in a fixed configuration and do not provide adequate decontamination of horizontal surfaces, which can be a major source of cross-contamination.  In the paper, we describe the design, implementation, and testing of an Ultraviolet Germicidal Irradiation (UVGI) system implemented on a mobile manipulation robot.  A human supervisor designates a surface for disinfection, the robot autonomously plans and executes an end-effector trajectory to disinfect the surface to the required certainty, and then displays the results for the human supervisor to verify.  We also provide some background information on UVGI and describe how we constructed and validated mathematical models of Ultraviolet (UV) radiation propagation and accumulation.  Finally, we describe our implementation on a Fetch mobile manipulation platform, and discuss how the practicalities of implementation on a real robot affect our models.


\end{abstract}

\section{Introduction}

Within the past ten years, harmful viruses, such as Ebola, MERS, and SARS-COV-2, have resulted in severe health hazards to healthcare facilities worldwide. Virus contact transmission presents risks for healthcare workers treating patients who infect surfaces through bodily secretions. Transmission can occur if a caregiver touches those surfaces then touches their mouth, eyes, or nose \cite{rewar2014transmission}. Liberia, for example, suffered 192 deaths, an eight percent reduction in their healthcare workforce due to workers contracting Ebola, while providing aid to individuals infected by the virus \cite{who2015ebola}. Disinfection of hazardous workspaces can alleviate the immediate dangers of contaminated surfaces.

Surface disinfection can play an important role in preventing virus spread; however, disinfecting a large healthcare facility periodically can be a laborious and mundane process. There is potential for a person to miss or insufficiently disinfect a surface due to fatigue. The disinfection task also presents the risk for a person to contract a virus, even while wearing gloves \cite{doebbeling1988removal}. Appropriate personal protective equipment can help reduce the chance of contracting a virus; however, additional health risks may arise. Healthcare providers in West Africa experienced heat stress due to double layer protective garments \cite{potter2015ebola}. An alternative approach is to use semi-autonomous robots. 

A robot's repeatable and accurate motions, as well as the ability to work in various environmental conditions, can provide consistent and effective surface disinfection. A method for combating infectious viruses is exposing surfaces to UV light. UVGI is a method that uses UV light in order to inactive a virus, which hinders the virus's ability to replicate inside a host cell, rendering it non-infectious \cite{kowalski2010ultraviolet}. UV disinfection robots have been deployed in various facilities and identified as a preventative measure of disease spread. Limited research has investigated robot coverage path planning for UV surface disinfection, specifically for mobile manipulator robots. This manuscript addresses robot path planning for UV disinfection by developing a benchmark algorithm that considers math model parameters and UV light distribution.


We begin by providing a brief overview of viruses and practical UV disinfection in Section \ref{background}. Previous UV robot devices and their limitations are discussed in Section \ref{related}. The system design and implementation of effective virus inactivation are discussed in Sections \ref{system_design} and \ref{implementation}, respectively. Section \ref{experiments} elaborates on the experiment techniques. The UV coverage results are presented in Section \ref{results}, followed by the discussion in Section \ref{discussion}. The implications of the results on UV robots and concluding remarks on future work is provided in Section \ref{future_work}.





\section{Background}
\label{background}
In this section we present an overview of a virus structure, in order to better understand how UV-based inactivation occurs.  We then briefly survey chemical, detergent, and UVGI inactivation techniques, and finally UVGI modeling, an essential component of the presented system design, is discussed in greater detail. 
 
\subsection{Virus Structure and Surface Transfer}

A virus genome is a molecule made of either ribonucleic acid (RNA) or deoxyribonucleic acid (DNA). The genome is protected by a coat of protein molecules, also known as a capsid, that helps the virus bind and invade a cell. The enveloped genome will instruct the host cell to replicate the viral genome and produce proteins to make new capsids. The newly-formed viruses escape the cell by punching holes through the cell's outer membrane then find new host cells to continue the virus replication process. A person who contracts a virus becomes a reservoir of virus particles, which can be released through coughing or sneezing. The tiny mucus droplets secreted from coughing or sneezing infect surfaces.

Mbithi et al. \cite{mbithi1992survival} investigated the interchange effectiveness of Hepatitis A from an inanimate surface to an animate one, such as a person's hand. The study found that surface contamination is a significant epidemiological threat in spreading infections. In optimal environmental conditions, viruses can stay active for several days on inanimate surfaces \cite{sagripanti2010persistence, paintsil2014hepatitis, sattar2000foodborne, otter2016transmission}. A virus's ability to transfer from inanimate surface to an animate one as well as its activation period on a surface demonstrates the importance of frequent disinfection.


\subsection{Inactivation Techniques}
Standard methods of virus inactivation are chemical reaction (biocidal agents), detergents, and UV irradiation. Chemicals, such as hydrogen peroxide and sodium hypochlorite, are used to oxidize and destroy a virus's protein capsid exposing the virus's nucleic acids to the biocidal agents, resulting in inactivation \cite{kampf2020persistence}. The use of chemicals on surfaces can reduce virus transmission, but the chemicals can potentially damage tools or measuring equipment. 

Detergents break down the outer protein molecules, damaging the virus's overall structure and inactivating the virus. Although effective, detergents do not work immediately and require thorough cleaning, which becomes impractical as facilities become larger in size. 

UVGI inactivates viruses by causing photochemical changes in the nucleic acids of the virus. A covalent bond between adjacent thymine (in DNA) or adjacent uracil (in RNA) molecules are formed, which leads to mutation and terminate replication \cite{kowalski2010ultraviolet, kowalski2009genomic}.

\subsection{UVGI Modeling}
UV light wavelengths between 200 and 290 nanometers (\(nm\)) have the characteristics of being germicidal \cite{kowalski2010ultraviolet,kesavan2012disinfection, nyangaresi2018effects}. The most effective wavelength range for germicidal irradiation is between 250-265 \(nm\) \cite{kowalski2010ultraviolet, kesavan2012disinfection}, which is within the UV-C band (200 - 280 \(nm\)).


Viruses are more resilient on a surface than in the air \cite{king2011germicidal, kowalski2010ultraviolet}, which causes an increase of resistance and challenges researchers to make accurate mathematical models of surface disinfection. The database for UV rate constants for airborne viruses is an adequate estimate for contaminated surfaces \cite{kowalski2010ultraviolet}. 

Virus populations decay exponentially when exposed to UV light \cite{kowalski2010ultraviolet, arguellesestimating}. The first-order decay rate model for UV irradiation is expressed as 

\begin{equation}\label{eq:1}
S = e^{-kD} ,
\end{equation}

\noindent
where \(S \) is the survival fraction, which is the ratio of viruses before and after disinfection, \(k \) (\(m^2/J\)) is the UV rate constant, and \(D\) (\(J/m^2\)) is the UV exposure dose. The UV exposure dose is defined as 

\begin{equation} \label{eq:2}
D = E_t \cdot I_R ,
\end{equation}

\noindent
where \(I_r \) (\(W/m^2\)) is the irradiance, which is the radiative flux through a flat surface, and \(E_t \) (\(s\)) is the exposure time. A UV dose of 90\% disinfection rate (10\% Survival) is expressed as  \(D_{90} \). \(I_r \) is approximated by


\begin{equation}\label{eq:3}
I_r = \frac{\eta P}{A_{exposed}} ,
\end{equation}

\noindent
where \(A_{exposed}\) is the exposed surface area, P is the power rating of source, and \(\eta \) is the attenuation factor where \(\eta \leq 1 \). A conservative estimate of \(\eta\) is \( 0.1\)  \cite{arguellesestimating}.



\section{Related Work}
\label{related}

\begin{figure}[h]
    \centering
    \begin{subfigure}{.47\linewidth}
        \centering
        \includegraphics[width=\linewidth]{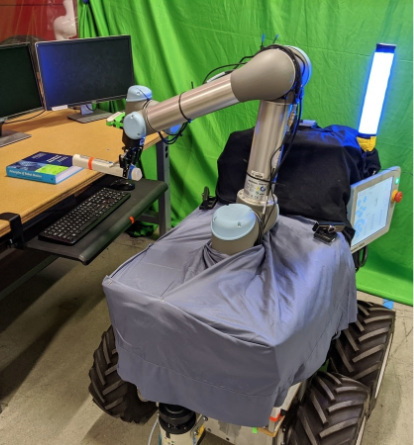}
        \caption{Agile Dexterous Autonomous Mobile Manipulation System Robot.}
        \label{fig:ADAMMS}
    \end{subfigure}
    \hspace{1mm}
    \begin{subfigure}{.47\linewidth}
        \centering
        \includegraphics[width=\linewidth]{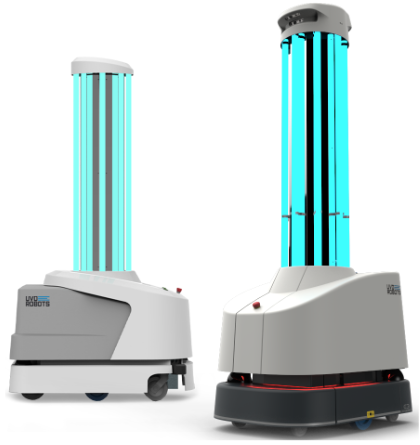}
        \caption{UV-Disinfection Robot.}
        \label{fig:UVD}
    \end{subfigure}
    \caption{UV-C robots}
    \label{fig:robots}
\end{figure}

I-Robot UV-C \cite{guettari2020uvc} and UV-Disinfection Robot \cite{rubaekevaluation}, identified in Fig. \ref{fig:UVD}, are common system designs for UVGI. The robots are designed with UV-C lamps around a central column, mounted on a mobile base. 
The design limitations associated with robots, such as the UV-Disinfection robot, is a poor line of sight from the UV-C lamps. Tabletops, a common virus transfer surface, are largely parallel to the light emitted and can receive relatively little direct illumination. Shadowing effects can also cause a UV-C mobile robot to inadvertently miss surfaces.
A method to address these limitations is to have the UV-C light source mounted on the arm of a mobile manipulator robot.

UV mobile manipulators have allowed robots to practice thorough UV disinfection of contaminated surfaces. The Agile Dexterous Autonomous Mobile Manipulation System (ADAMMS), Fig. \ref{fig:ADAMMS}, is a robot that maneuvers a UV wand across intricate shaped objects and non-planar surfaces for disinfection \cite{thakar2021usc}. The different poses of the wand ensure all angles of a surface are considered. Although this system design is a step forward in UV disinfection robots, little work has investigated mapping and quantifying the UV dose from a planned path. This manuscript seeks to improve the efficacy of disinfection is systems, such as the ADAMMS, in order to validate the UVGI models to the UV coverage planner.



\section{System Design}
\label{system_design}

The system design involves a mobile manipulation robot that sweeps an ultraviolet light source over infected surfaces.  A human supervisor designates the parts of the surface to be disinfected, and specifies the disinfection rate.  Armed with a 3-dimensional (3D) model of the world built from sensor data, the surface to be disinfected, and the disinfection rate, we plan a path and speed for the robot end effector that will accomplish the task.  The plan is executed by the robot, and a visualization of the UV light coverage is presented back to the supervisor for approval.

The plan first builds an approximate, voxel-based model of the world.  The model is then presented to the human supervisor (see Fig. \ref{fig:rviz}), who can place and move markers directly into the 3D visualization to define the extent of the area to be disinfected.  A plane is fit to the markers, and the plan of robot motion is developed with respect to the plane.

A plan using a lawnmower coverage pattern over the plane, and calculate an appropriate speed to ensure that every point on the place receives enough UV radiation to inactivate the virus.  This is done by projecting the plane down onto the 3D model, and taking the distance from the UV source to the actual surface into account. As the robot moves along the planned path, the actual motions update a discretized accumulator array that tracks the UV radiation delivered, based on Equation \ref{eq:2} and an assumption of a perpendicular projection of light and a circular area of illumination.

The accumulator array is presented to the human supervisor in the visual interactive interface allowing them to determine if a sufficient amount of UV light has been delivered, according to the models and the actual motion of the robot.  The visual interactive interface can show the accumulated amount using increasing opacity (Fig. \ref{fig:heatmaps}).

\section{Implementation}
\label{implementation}

An overview of the implementation of the system design, including the manipulator robot physical capabilities and how the visual interactive interface enabled the human supervisor to control the system, are discussed. How the UV path planner computed the required manipulator velocity is also provided. 

 \begin{figure}[h]
 \centering
 \includegraphics[width=.75\columnwidth]{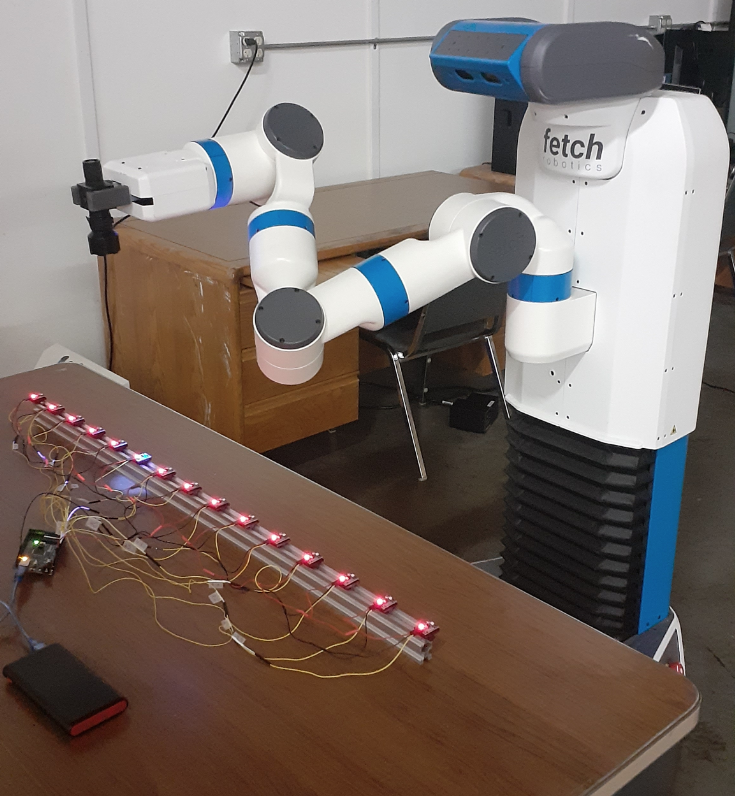}
 \caption{A Fetch robot hovering UV flashlight above UV sensor array.}
 \label{fig:Fetch_robot}
 \end{figure}

\subsection{Robot System}
A Fetch mobile manipulator, shown in Fig. 2, was used to test the automated UV coverage planner. Fetch has a 7 degrees of freedom arm that can carry a maximum payload of 6 \(kg\) at full extension and operate at a maximum velocity of 1 \(m/s\) \cite{wise2016fetch}. Fetch is equipped with a Primesense Carmine short-range depth camera, which was used for the visual interactive interface.  

Fetch was designed to work with the open-source robot software ecosystem, Robot Operating System (ROS) \cite{quigley2009ros}, which was used to control the manipulator, model the UV light distribution, automate path planning, and allow human supervision during disinfection.

\subsection{User Interface}
\begin{figure}[h]
    \centering
    \begin{subfigure}{.75\linewidth}
        \centering
        \includegraphics[width=\linewidth]{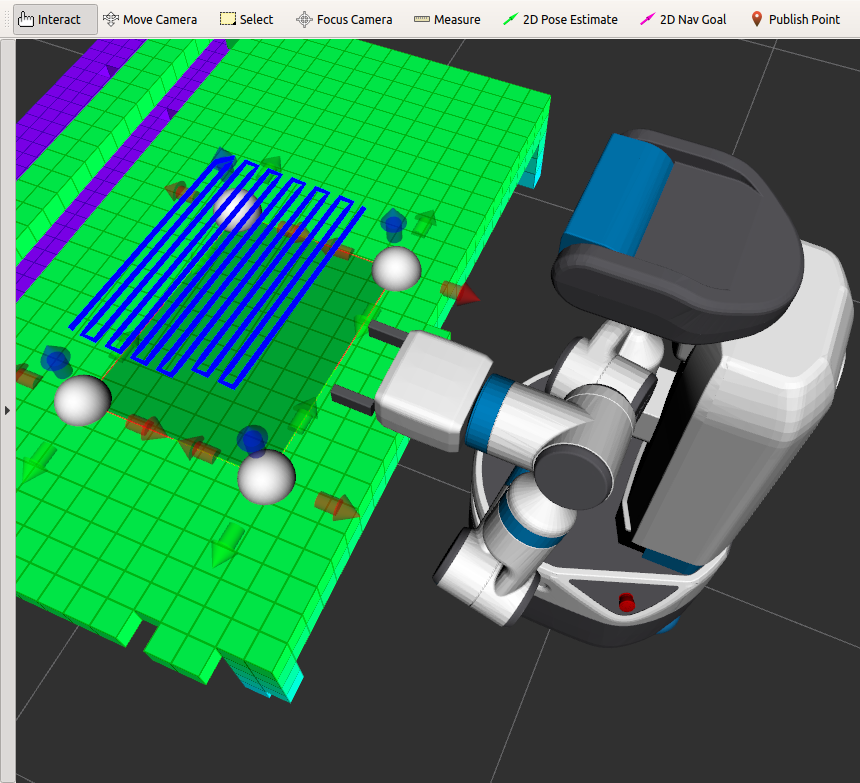}
        \caption{Rviz interface.}
        \label{fig:rviz}
    \end{subfigure}
    \vspace{3mm}
    \begin{subfigure}{.75\linewidth}
        \centering
        \includegraphics[width=\linewidth]{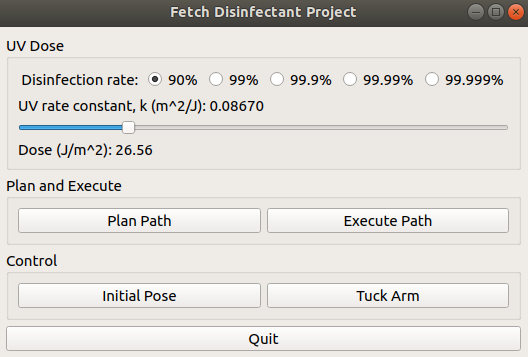}
        \caption{Robot control interface.}
        \label{fig:gui}
    \end{subfigure}
    \caption{Images of the rviz (a) and menu (b) interfaces.}
    \label{fig:interfaces}
\end{figure}

The scene in rviz, the ROS visualization tool, is shown in Fig. \ref{fig:rviz}. The human supervisor can designation 3D points in the scene with mouse clicks, to define a region of interest. These points (which are represented using rviz interactive markers) can then be adjusted, as needed, and represent the vertices of the disinfection region, which is represented by a semi-transparent plane. The blue line strip is the path that the robot's end effector will follow.

The interface, shown in Fig. \ref{fig:gui}, displays a menu to input the parameters for the first order decay model. The \(k\) value is set with a linear slider. The disinfection rate ranges logarithmically from 90\% to 99.999\%, following conventions in the infectious disease community, and is selected with a radio button. The required UV dose is displayed for reference when the parameters are modified. The ``plan path'' and ``execute path'' buttons cause the robot to plan and execute, respectively, a path that is calculated based on the parameters set in this interface.

\subsection{Path Planning}
Three parameters are considered to approximate the manipulator velocity: 1) the disinfection rate, 2) the \(k\) value, 3) and the UV light distribution. The disinfection rate and \(k\) value are set from the GUI inputs. The UV light distribution is defined as a N by N matrix and is referred to as a kernel mask. Each element in the kernel mask is associated with a physical size and is defined as 

\begin{equation} \label{eq:4}
e_{size} = \frac{d_{exposed}}{N_{size}} ,
\end{equation}

\noindent
where \(e_{size}\) (\(m\)) is the height and width of the element, \(d_{exposed}\) (\(m\)) is the diameter of the UV exposed area, and \(N_{size}\) is the number of rows or columns of the N by N matrix. The UV exposure time for an element is expressed as

\begin{equation} \label{eq:5}
E_t = \frac{e_{size}}{V_{max}} ,
\end{equation}

\noindent
where \(V_{max}\) (\(m/s\)) is the maximum velocity of the manipulator. The exposure time and irradiance values across the center row of the kernel mask, denoted as \(arr_{mask}\), are used as inputs for a sum function (Algorithm \ref{alg:loop}).

\begin{algorithm}[H] 
\caption{Sum of dose values}
\label{alg:loop}
\begin{algorithmic}[1]
\Require{$E_{t} , arr_{mask}$} 
\Ensure{$D_{min}$ (sum of dose)}
\Statex
\Function{Sum}{$E_{t} , arr_{mask}$}
  \State {$D_{min}$ $\gets$ {$0$}}
  \State {$L$ $\gets$ {$length(arr_{mask})$}}
   \For{$k \gets 1$ to $L$}                    
       \State {$D_{min}$ $\gets$ {$D_{min} + arr_{mask}[k] \times E_{t}$}}
   \EndFor
   \State \Return {$D_{min}$}
\EndFunction
\end{algorithmic}
\end{algorithm}

The sum function returns \(D_{min}\), which represents the minimum dose an output element will experience at maximum manipulator velocity. Dividing the minimum dose by the menu's computed required dose results in the scaling velocity factor. The equation is

\begin{equation}\label{eq:7}
SF = \frac{D_{min}}{D_{req}} ,
\end{equation}

\noindent
where \(SF\) is the scaling velocity factor, and \(D_{req}\) is the required dose. \(SF\) is then subscribed to the coverage planner.

\section{Experiments}
\label{experiments}
Experiments were performed to confirm that the system design can model and measure UV dose accurately. The validation experiments section discusses UV light distribution model and how the model was represented in the coverage planner and rviz. The system experiments provide information on the performance and tolerances of the UV measuring apparatus.


\begin{figure}[h]
 \centering
 \includegraphics[width=.5\columnwidth]{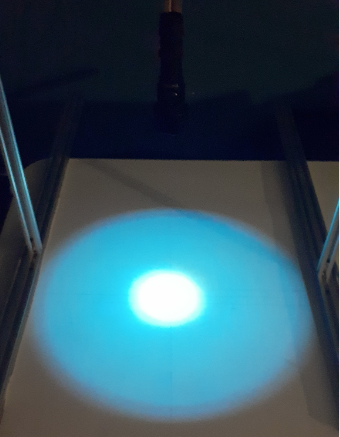}
 \caption{ Light distribution emitted from UV Flashlight.}
 \label{fig:light_distribution}
 \end{figure}

\subsection{Validation Experiments}
The UV light distribution was defined using the two-dimensional kernel mask. The employed UV light source was the 10 Watt TATTU U2S UV Flashlight, which emitted UV light at 365 \(nm\). The flashlight was mounted approximately 0.3 \(m\) above a General Tools Digital UV-A and UV-B meter. The UV digital meter's calibration point is 365 \(nm\) and has a measuring accuracy of \(\pm\) 4\% of the reading. Irradiance was measured at increasing distances from the center of the UV exposed area. Fig. \ref{fig:light_distribution} illustrates the non-uniform light distribution. 

\begin{figure}[h]
 \centering
 \includegraphics[width=.9\columnwidth]{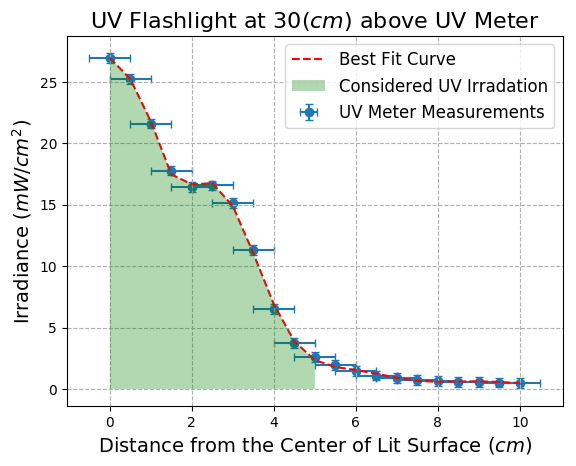}
 \caption{ Distance vs Irradiance.}
 \label{fig:D_vs_I}
 \end{figure}

A 15th order polynomial was fit to the dataset, shown in Fig. \ref{fig:D_vs_I}. The bump in the graph at 2.5 \(cm\) is due to the UV LED's circular configuration, which emitted hot spots and rings. The polynomial equation is then broadcasted to the path planner for further computations. Irradiance was measured up to 16 \(cm\) away from the center of the lit surface, however, the intensity quickly drops to negligible values past 8 \(cm\).

\begin{figure}[h]
 \centering
 \includegraphics[width=.95\columnwidth]{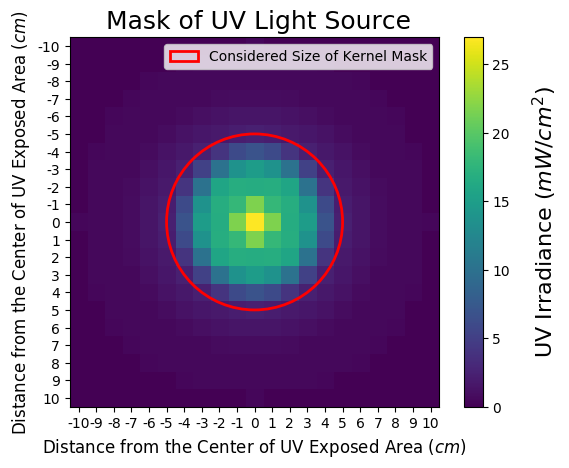}
 \caption{ Ramped Mask Colorbar.}
 \label{fig:ramped_2}
 \end{figure}

The path planner used the polynomial equation to input UV irradiance values into the kernel mask, which represented the UV flashlight distribution, shown in Fig. \ref{fig:ramped_2}. The kernel mask is spanned across an output surface of a larger two-dimensional array. This spanning technique can model theoretical dose values for a 0.1 \(m\) by 1 \(m\) output surface.


\begin{figure}[h]
    \centering
    \begin{subfigure}[t]{.47\linewidth}
        \centering
        \includegraphics[width=\linewidth]{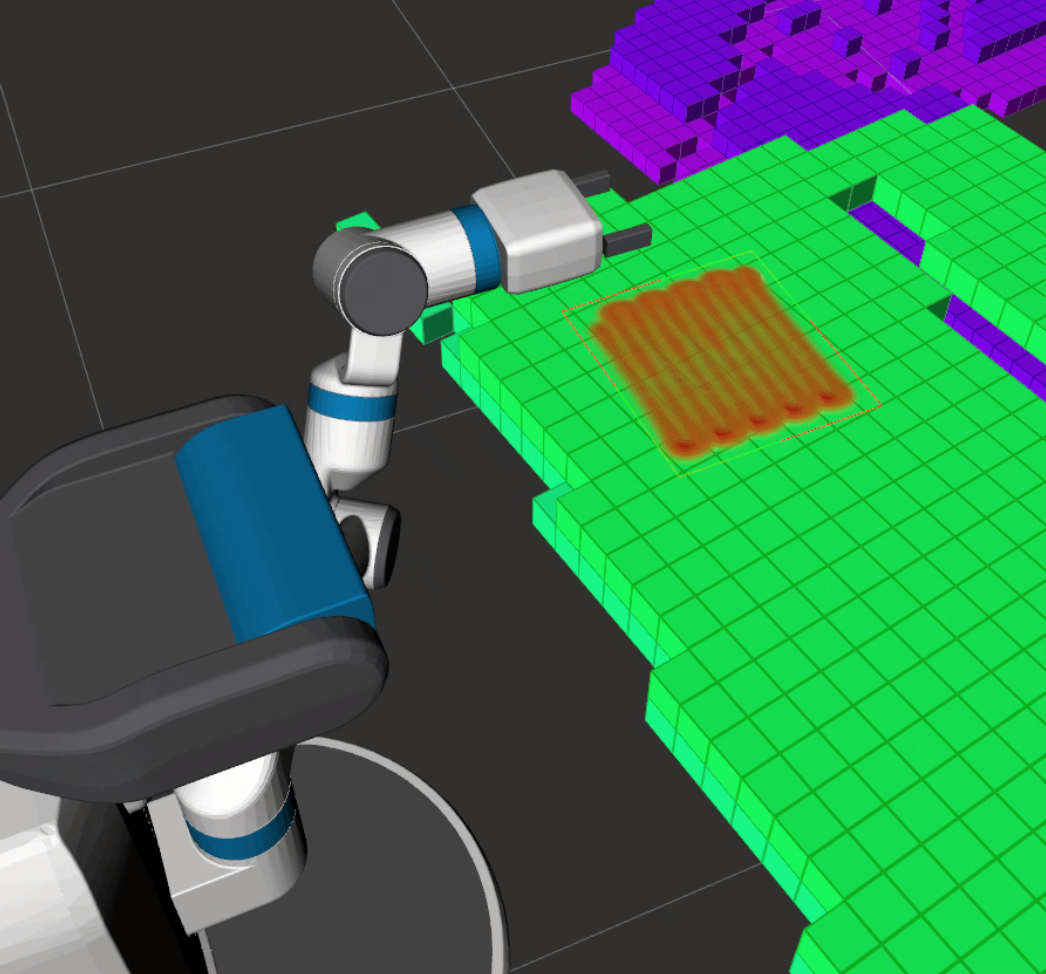}
        \caption{Isometric view of heat map visuals.}
        \label{fig:heatmap_3}
    \end{subfigure}
    \hspace{1mm}
    \begin{subfigure}[t]{.49\linewidth}
        \centering
        \includegraphics[width=\linewidth]{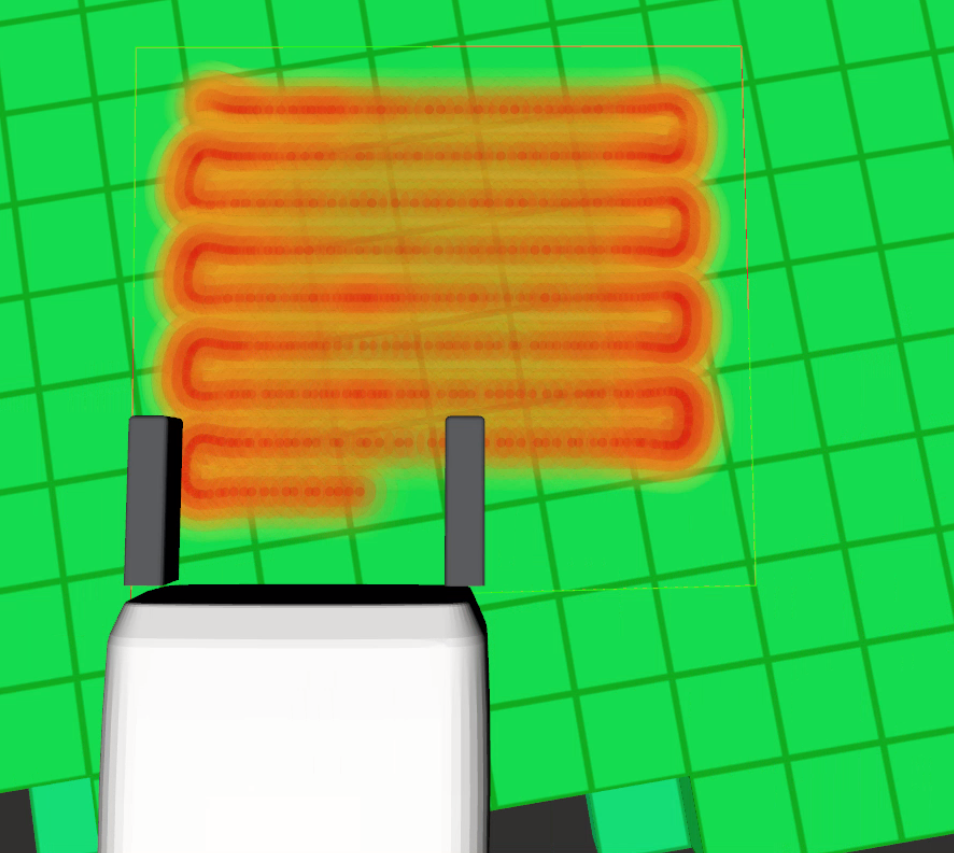}
        \caption{Top view of heat map visuals.}
        \label{fig:heatmap_4}
    \end{subfigure}
    \caption{Different views of heat map representation of irradiance distribution.}
    \label{fig:heatmaps}
\end{figure}

The path planner visualizes the irradiance of UV light distribution in the disinfection region in rviz. Both Fig. \ref{fig:heatmap_3} and Fig. \ref{fig:heatmap_4} show the irradiance intensity in the disinfection region with a heat map marker.

\subsection{System Experiments}
The measuring apparatus comprised of 15 Waveshare UV sensor modules. The Waveshare sensor's output voltage is linearly proportional to UV light intensity and is equipped with an onboard amplifier circuit to adjust the output signal's amplification. The sensors were calibrated with the UV digital meter and capable of measuring irradiance values up to 40 \(mW/cm^2\) with a tolerance of \(\pm\) 1 \(mW/cm^2\). Fig. \ref{fig:Fetch_robot} shows a set of UV modules installed on a 3D printed adapter and integrated on a 1 \(m\) long aluminum extrusion. All sensors were evenly spaced from one another. 

An Arduino Mega2560 was used to process the voltage output signals of the UV sensor array. The accumulation algorithm, conversions, and UVGI equations were programmed in the microcontroller. The Arduino Mega was connected via USB to our laptop, in which the measured dose of the sensors was exported as a comma separated value file.

\begin{figure}
 \centering
 \includegraphics[width=.9\columnwidth]{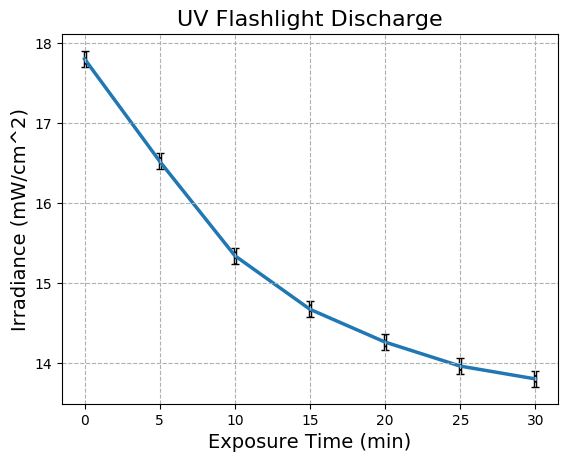}
 \caption{Power drop of UV flashlight as a function of time.}
 \label{fig:flashlight_discharge}
 \end{figure}

A rechargeable UV flashlight was connected to a portable power bank to keep a constant power rating during disinfection. The UV flashlight was held 0.4 \(m\) directly above the UV digital meter and measured for 30 \(min\). Fig. \ref{fig:flashlight_discharge} shows the UV flashlights time and irradiance relationship.


\section{Results}
\label{results}

\subsection{Model Results}
\begin{figure}[h]
 \centering
 \includegraphics[width=\columnwidth]{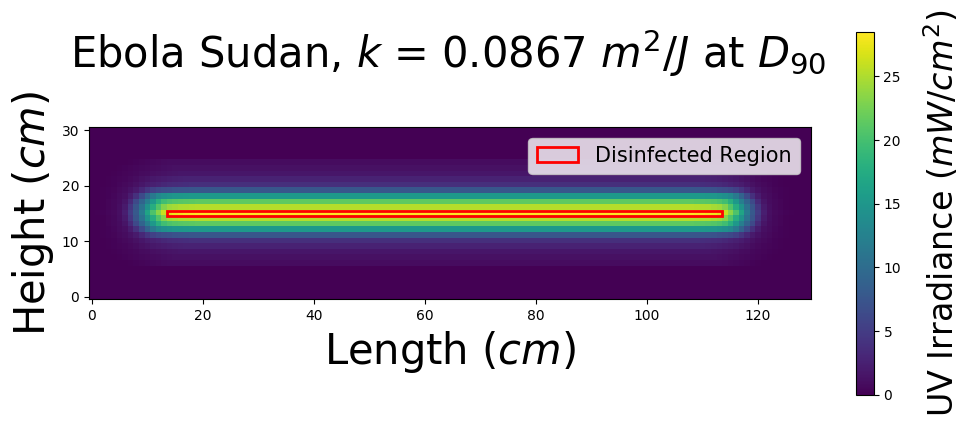}
 \caption{ 1-meter UV Coverage for Ebola Sudan.}
 \label{fig:coverage}
 \end{figure}

A model of the spanning technique was conducted for a \(k\) value of 0.0867 \(m^2/J\) at \(D_{90}\). The value represents the Ebola Sudan virus. Fig. \ref{fig:coverage} shows the distributed dose for a 1 \(m\) coverage path. The red highlighted disinfected region is a 1 by 100 array. The array corresponds to a size of 0.1 \(m\) by 1 \(m\). The disinfected region ranged from 27.3 to 28.5 \(J/m^2\), which is above the required dose of 27 \(J/m^2\). The \(k\) value at \(D_{90}\) approximates the translational arm velocity at 0.57 \(m/s\).  

\subsection{Physical Results}

\begin{figure}[h]
 \centering
 \includegraphics[width=.9\columnwidth]{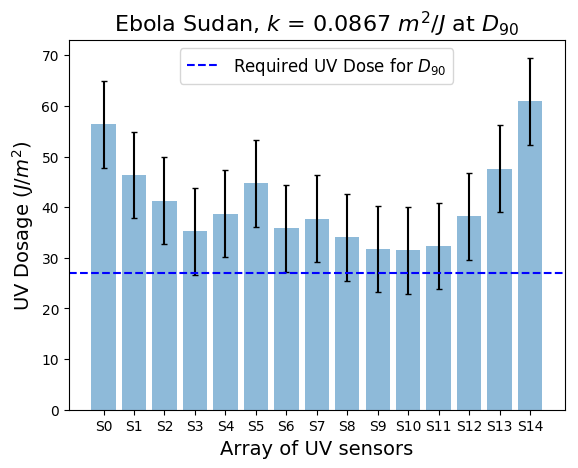}
 \caption{UV Dose measurements for Ebola Sudan at \(D_{90} \)}
 \label{fig:ebola_d90}
 \end{figure}

Fig. \ref{fig:ebola_d90} contains the measured dose for the 15 sensor array.  The error bars represent one standard deviation of the dose data. Each of the 15 sensors measured above the required UV dose of 27 \(J/m^2\). Sensors S3, S8, S9, S10, and S11 contained the required dose within the standard deviation. 

\begin{figure}
 \centering
 \includegraphics[width=\columnwidth]{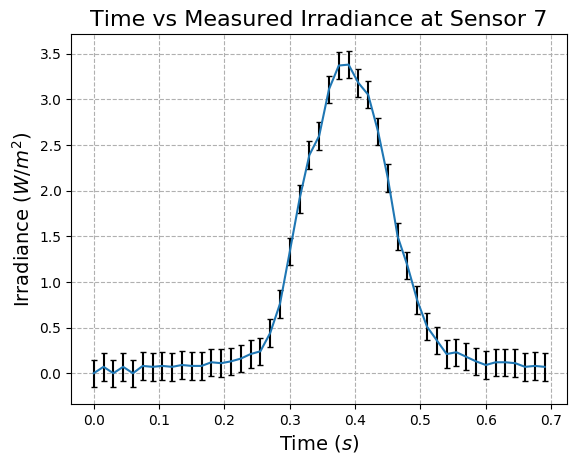}
 \caption{ UV Dose measurements from sensor S7.}
 \label{fig:ebola_s7}
 \end{figure}

Sensor S7's time versus irradiance is demonstrated in Fig. \ref{fig:ebola_s7}. Sensor S7 underwent UV irradiance for a total time of 0.69 \(s\). The sensor modules are capable of measuring irradiance up to 0.15 \(m\) away from the center of the light. The UV flashlight traveled approximately 0.3 \(m\) within 0.69 \(s\), which equates to 0.435 \(m/s\). The calculated velocity is 76.28\% of the targeted value of 0.57 \(m/s\). The dose for sensor S7 was measured at 37.7 \(J/m^2\).

\begin{figure}
 \centering
 \includegraphics[width=\columnwidth]{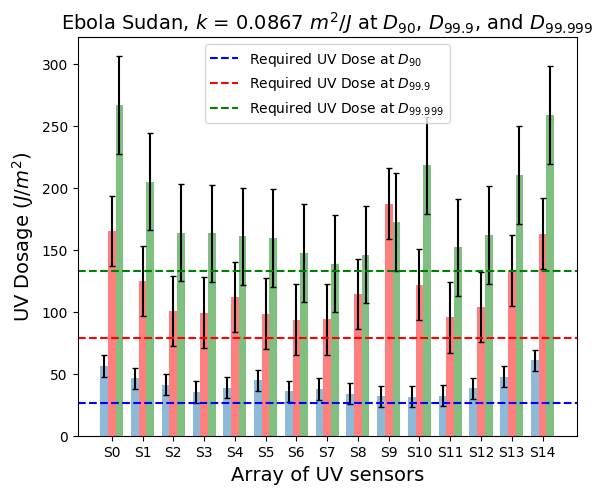}
 \caption{UV Dose measurements for Ebola Sudan for three different disinfection dosages. }
 \label{fig:ebola_diff_d}
 \end{figure}

Fig. \ref{fig:ebola_diff_d} contains UV dose measurements for the Ebola virus at \(D_{90} \), \(D_{99.9} \), and \(D_{99.999}\). The error bars represent one standard deviation with the respective disinfection rate data. The results of \(D_{90} \) were previously mentioned. 

The required dose for \(D_{99.9} \) is 79.65 \(J/m^2\) with the arm velocity set at 0.19 \(m/s\). Each sensor measured above the required dose for \(D_{99.9} \). 7 out of the 15 UV sensor modules contained the required dose within the standard deviation.

\(D_{99.999} \) requires a dose of 132.74 \(J/m^2\) with arm velocity set at 0.11 \(m/s\). Each sensor measured above the required dose for \(D_{99.999} \). Sensors S0, S1, S10, S13, and S14, had the standard deviation error bars above the required dose. 

\section{Discussion}
\label{discussion}
The results indicate that the parametrizations rendered sufficient UV dose levels for both the model and physical tests. The disinfection region in the output surface stayed within a consistent range above the needed dose, demonstrating that the model performed as needed. The Fetch robot provided the required UV dose across the 1 \(m\) sensor array, which validates the UV path planner. The developed UV path planner can further develop UV disinfection robots because the planner can quantify and provide the required UV dose for various viruses. While the UV coverage planner worked for several cases, we did encounter limitations within the design. 

The dynamics of the manipulator caused a dose saturation on several UV sensor modules. The manipulator consistently accelerated and reached the set velocity halfway through the sensor array. The first few sensors experienced a larger exposure time due to the slower arm velocity, which resulted in larger dose values. The same behavior was observed for the last set of sensors when the manipulator was decelerating to stop at the last waypoint. 

The UV emission of the flashlight was another limitation. The irradiance dropped approximately 2.46 \(W/m^2\) after ten minutes of use. The temperature of the flashlight increased during the discharge experiment. The heightened temperature may have increase the flashlight's internal electronics' resistivity and resistance, which can result in a lower power output. 

Our current implementation depends on an assumption of planar surfaces.  This means that any objects on top of this surface will receive more irradiation.  The surplus of irradiation is not bad, since it will increase the chances that any virus present will be inactivated.  However, any areas below the surface will receive less irradiation, which increases the chances that any virus present will not be inactivated.  Currently, we address this by having the human supervisor place the plane at the lowest point in the region.





\section{Future Work}
\label{future_work}
There are several areas for improvement to our UV coverage planning approach. We can enhance the accumulation algorithm to work directly with the 3D model of the world (currently implemented using an oct-tree) rather than 2D planes. This would allow the system to model coverage in a 3D workspace, and to ensure complete irradiation regardless of the geometry of the surface. Another improvement is to implement adaptive modeling of the UV light distribution. For instance, using an on-board UV sensor that we can periodically illuminate with our UV light to more accurately model the source's current power output. This would allow the algorithm to adaptively modify the end-effector speed to compensate for the drop in power. We can also refine the model to reflect the dynamics of the physical robot more accurately to provide a more consistent dose across infected surfaces. 




\bibliographystyle{IEEEtran}
\bibliography{refs}

\end{document}